\documentclass{svproc}
\usepackage{graphicx}
\usepackage{longtable}
\begin{document}
\title{g2tmn at Constraint@AAAI2021: Exploiting CT-BERT and Ensembling Learning for COVID-19 Fake News Detection}
%
\titlerunning{CT-BERT and Ensembling Learning for COVID-19 Fake News Detection}
%

\author{Anna Glazkova\inst{1}\unskip$^{[{0000-0001-8409-6457}]}$ \and
Maksim Glazkov\inst{2}\unskip$^{[{0000-0002-4290-2059}]}$ \and
Timofey Trifonov\inst{1,2}\unskip$^{[{0000-0001-7996-0044}]}$}

\authorrunning{A. Glazkova et al.} 
%
\tocauthor{Anna Glazkova, Maksim Glazkov, and Timofey Trifonov}
\institute{University of Tyumen, ul. Volodarskogo 6, 625003 Tyumen, Russia \email{a.v.glazkova@utmn.ru}, \email{timak474@gmail.com}\and
"Organization of cognitive associative systems" LLC, ul. Gertsena 64, 625000 Tyumen, Russia
\email{my.eye.off@gmail.com}}
\maketitle              
\begin{abstract}
The COVID-19 pandemic has had a huge impact on various areas of human life. Hence, the coronavirus pandemic and its consequences are being actively discussed on social media. However, not all social media posts are truthful. Many of them spread fake news that cause panic among readers, misinform people and thus exacerbate the effect of the pandemic. In this paper, we present our results at the Constraint@AAAI2021 Shared Task: COVID-19 Fake News Detection in English. In particular, we propose our approach using the transformer-based ensemble of COVID-Twitter-BERT (CT-BERT) models. We describe the models used, the ways of text preprocessing and adding extra data. As a result, our best model achieved the weighted F1-score of 98.69 on the test set (the first place in the leaderboard) of this shared task that attracted 166 submitted teams in total.

\end{abstract}
\section{Introduction}

Social media is a unique source of information. On the one hand, their low cost, easy access and distribution speed make it possible to quickly share the news. On the other hand, the quality and reliability of social media news is difficult to verify \cite{intr}. This is the source of a lot of false information that has a negative impact on society. 

Over the past year, the world has been watching the situation developing around the novel coronavirus pandemic. The COVID-19 pandemic has become a significant newsworthy event of 2020. Therefore, news related to COVID-19 are actively discussed on social media and this topic generates a lot of misinformation. Fake news related to the pandemic have large-scale negative social consequences, they provoke huge public rumor spreading and misunderstanding about the COVID-19 and aggravate effects of the pandemic. Moreover, recent studies \cite{med1} show an increase in symptoms such as anxiety and depression in connection with the pandemic. This is closely related to the spread of misinformation, because fake news can be more successful when the population is experiencing a stressful psychological situation \cite{med2}. The popularity of fake news on social media can rapidly increase, because the rebuttal is always published too late. In this regard, there is evidence that the development of tools for automatic COVID-19 fake news detection plays a crucial role in the regulation of information flows.

In this paper, we present our approach for the Constraint@AAAI2021 Shared Task: COVID-19 Fake News Detection in English \cite{overview} that attracted 433 participants on CodaLab. This approach achieved the weighted F1-score of 98.69 (the first place in the leaderboard) on the test set among 166 submitted teams in total.

The rest of the paper is organized as follows. A brief review of related work is given in Section 2. The definition of the task has been summarized in Section 3, followed by a brief description of the data used in Section 4. The proposed methods and experimental settings have been elaborated in Section 5. Section 6 contains the results and error analysis respectively. Section 7 is a conclusion.

\section{Related Work}

In recent years, the task of detecting fake news and rumors is extremely relevant. False information spreading involves various research tasks, including: fact checking \cite{checkthat,fever1}, topic credibility \cite{kim,fever}, fake news spreaders profiling \cite{spreaders}, and manipulation techniques detection \cite{prop}. Various technologies and approaches in this field range from traditional machine learning methods \cite{buda,utmn,pizarro}, to state-of-the-art transformers \cite{morio,wu}.

A overview of fake news detection approaches and challenges on social media has been discussed in \cite{intr,review}. Many scholars have proposed their solutions to this problem in different subject areas (in particular, \cite{rel2,rel1}). Up to now, a large number of studies in fake news detection have used supervised methods including models based on transformers-based architecture \cite{rel4,rel5,rel3}. 

Some recent work have focused on detecting fake news about COVID-19. For example, the predictors of the sharing of false information about the pandemic are discussed in \cite{sharing}. In \cite{albert}, a novel COVID-19 fact checking algorithm is proposed that retrieves the most relevant facts concerning user claims about particular facts. A number of studies have begun to examine COVID-19 fake news detection methods for non-English languages \cite{preprint1,preprint,preprint2}.

In addition, several competitions have been announced over the past year related to the analysis of posts about COVID-19 on social media \cite{st2,st1,competition}. 

\section{Task Definition}

The task focused on the detection of COVID-19-related fake news in English. The sources of data was various social-media platforms such as Twitter, Facebook, Instagram, etc. Formally, the task is described as follows.
\begin{itemize}
    \item \textbf{Input.} Given a social media post.
    \item \textbf{Output.} One of two different labels, such as "fake" or "real".
\end{itemize}

The official competition metric was F1-score averaged across the classes (the weighted F1-score). The participants were allowed five submissions per team throughout the test phase.

\section{Dataset}

The dataset \cite{dataset} provided to the participants of the shared task contains 10,700 manually annotated social media posts divided into training (6420), validation (2140), and test (2140) sets. The vocabulary size (i.e., unique words) of the dataset is 37,505 with 5141 common words in both fake and real news. The dataset contains the post ID, the post, and the corresponding label which is "fake" or "real" (see Table 1).

\begin{longtable}{|p{0.15\linewidth}|p{0.85\linewidth}|}
\caption{Some examples of fake and real posts}
\\
\hline
Label & Post \\ \hline
real  & The CDC currently reports 99031 deaths. In general the discrepancies in death counts between different sources are small and explicable. The death toll stands at roughly 100000 people today \\ \hline
real  & \#IndiaFightsCorona: We have 1524 \#COVID testing laboratories in India and as on 25th August 2020 36827520 tests have been done : @ProfBhargava DG @ICMRDELHI \#StaySafe \#IndiaWillWin https://t.co/Yh3ZxknnhZ \\ \hline
fake & Politically Correct Woman (Almost) Uses Pandemic as Excuse Not to Reuse Plastic Bag https://t.co/thF8GuNFPe \#coronavirus \#nashville \\ \hline
fake & Obama Calls Trump’s Coronavirus Response A Chaotic Disaster https://t.co/DeDqZEhAsB \\ \hline
\end{longtable}

\section{Our Approach}

In this section, we describe the approaches that we evaluated on the validation data during the validation phase. We used transformer-based models as they demonstrate state-of-the-art results in most text classification tasks. We also evaluated the empirical effectiveness of a Linear Support Vector baseline (Linear SVC) and different text preprocessing techniques and adding extra data. The results are shown in the next section.

\subsection{Data Preprocessing}

Our approaches to text preprocessing for transformer-based models are various combinations of the following steps, most of which have been inspired by \cite{insp1,insp2}:

\begin{itemize}
    \item removing or tokenizing hashtags, URLs, emoji, and mentions using a preprocessing library for tweet data written in Python \cite{tp}. Tokenization means the replacement of URLs, mentions, and emoji with special tokens, such as \$URL\$, \$MENTION\$, and \$HASHTAG\$ respectively (for example "HHS to distribute \$4 billion to \#COVID hot spots; \$340 million already paid out. https://t.co/uAj29XA1Y5" (original) $\rightarrow$ "HHS to distribute \$4 billion to \$HASHTAG\$ hot spots; \$340 million already paid out. \$URL\$" (tokenizing); "HHS to distribute \$4 billion to hot spots; \$340 million already paid out." (removing)); 
    \item using the Python emoji library to replace the emoji with short textual description \cite{em}, for example :red\_heart:, :thumbs\_up:, etc.;
    \item converting hashtags to words ("\#COVID" $\rightarrow$ "COVID");
    \item translating in the lowercase.
\end{itemize}

In the case of the baseline, we translated the text to the lowercase, removed punctuation and special characters, and then lemmatized the words. Further, we converted texts into the form of a token counts matrix (a bag of words model).

\subsection{Models}

We experimented with the following transformer-based models:

\begin{itemize}
    \item \textbf{BERT} \cite{bert}. BERT is a language representation model presented by Google, which stands for Bidirectional Encoder Representations from Transformers. BERT-based models show great results in many natural language processing tasks. In our work, we used BERT-base-uncased, which is pretrained on texts from Wikipedia.
    \item \textbf{RoBERTa} \cite{roberta}. RoBERTa is a robustly optimized BERT approach introduced at Facebook. Unlike BERT, RoBERTa removes the Next Sentence Prediction task from the pretraining process. RoBERTa also uses larger batch sizes and dynamic masking so that the masked token changes while training instead of the static masking pattern used in BERT. We experimented with RoBERTa-large.
    \item \textbf{COVID-Twitter-BERT} \cite{ctb}. CT-BERT is a transformer-based model, pretrained on a large corpus of Twitter messages on the topic of COVID-19 collected during the period from January 12 to April 16, 2020. CT-BERT is optimised to be used on COVID-19 content, in particular social media posts from Twitter. This model showed a 10–30\% marginal improvement compared to its base model, BERT-large, on five different specialised datasets. Moreover, it was successfully used for a variety of natural language tasks, such as identification of informative COVID-19 tweets \cite{insp1}, sentiment analysis \cite{sent}, and claims verification \cite{claim1,claim2}.
\end{itemize}

\subsection{Additional Data}

To improve the quality of our approach, we made attempts to add extra data to the model. For this purpose we used two datasets related to the topic of COVID-19 fake news:

\begin{itemize}
    \item \textbf{CoAID: COVID-19 Healthcare Misinformation Dataset} \cite{add_data2}. The dataset includes 4251 health-related fake news posted on websites and social platforms. 
    \item \textbf{FakeCovid - A Multilingual Cross-domain Fact Check News Dataset for COVID-19} \cite{add_data1}. The dataset contains 5182 fact-checked news articles for COVID-19 collected from January to May 2020. 
\end{itemize}

In our experiments, we added news headlines to the training set.

\subsection{Experimental Settings}

We conducted our experiments on Google Colab Pro (CPU: Intel(R) Xeon(R) CPU @ 2.20GHz; RAM: 25.51 GB; GPU: Tesla P100-PCIE-16GB with CUDA 10.1). Each model was trained on the training set for 3 epochs and evaluated on the validation set. As our resources are constrained, we used random seeds to fine-tune pre-trained language models and made attempts to combine them with other parameters. The models are optimised using AdamW \cite{adam} with a learning rate of 2e-5 and epsilon of 1e-8, max sequence length of 128 tokens, and a batch size of 8. We implemented our models using Pytorch \cite{pytorch} and Huggingface’s Transformers \cite{hft} libraries. 

The Linear SVC was implemented with Scikit-learn \cite{sklearn}. For text preprocessing, we used NLTK \cite{nltk} and Scikit-learn's CountVectorizer with a built-in list of English stop-words and a maximum feature count of 10,000.

\section{Results and Discussion}

\subsection{Comparison of Models for Fake News Detection}

In Table 2, we present the results of our experiments in a step by step manner. We started with a Linear SVC baseline and then evaluated BERT-based models using a variety of text preprocessing and adding extra data techniques. Note that we evaluated our models using F1-score for the fake class while the official competition metric was the weighted F1-score.

\begin{longtable}{|p{0.20\linewidth}|p{0.40\linewidth}|p{0.15\linewidth}|p{0.15\linewidth}|}
\caption{Evaluation results}
\\
\hline
Model & Data preprocessing & Additional data & F1-Score (\%, for fake class) \\ \hline
LinearSVC & converting into a bag of words & no & 88.39 \\ \hline
BERT & lowercase & no & 96.75 \\ \hline
RoBERTa & lowercase & no & 97.62 \\ \hline
RoBERTa & removing hashtags, URLs, emoji + lowercase & no & 95.79 \\ \hline
RoBERTa & removing URLs and emoji + converting hashtags to words + lowercase & no & 95.68 \\ \hline
CT-BERT & lowercase & no & 98.07 \\ \hline
CT-BERT & tokenizing URLs and mentions + converting emoji to words + lowercase & no & 97.87 \\ \hline
CT-BERT & converting emoji to words + lowercase & no & 98.32 \\ \hline
CT-BERT & tokenizing URLs + converting emoji to words + lowercase & no & 98.42 \\ \hline
CT-BERT & tokenizing URLs + converting emoji to words + lowercase & FakeCovid & 98.23 \\ \hline
CT-BERT & tokenizing URLs + converting emoji to words + lowercase & CoAID & 98.37 \\ \hline
\end{longtable}

As can be seen from the table above, CT-BERT models showed absolutely better results compared to BERT- and RoBERTa-based classifiers. Our work doesn't contain a detailed comparative analysis of text preprocessing techniques for this task. Still we can see that text preprocessing can affect the quality of fake news detection. For example, there was no evidence that removing emoji and mentions improve the model results. A clear benefit of converting hashtags into words could not be identified during this evaluation. Also, as a result of our experiments, we decided not to tokenize links to other user's accounts (mentions). This can be seen in the case of mentions of major news channels like CNN that tend to indicate that the post is real. The next section of the model evaluation was concerned with using additional datasets. We noticed that adding extra data did not show any benefits in our experiments

\subsection{Final Submissions}

 As it was mentioned above, the participants of the shared task were allowed five submissions per team throughout the test phase. Our best model based on experimental results (subsection 5.1) included the following preprocessing steps: tokenizing URLs, converting emoji to words, and lowercase. As final submissions, we used the results of hard voting ensembles of three such models with random seed values and with different data splitting into training and validation samples. The final architecture of our solution is shown in Figure 1. 

\begin{figure}[h]
\center{\includegraphics[width=1\linewidth]{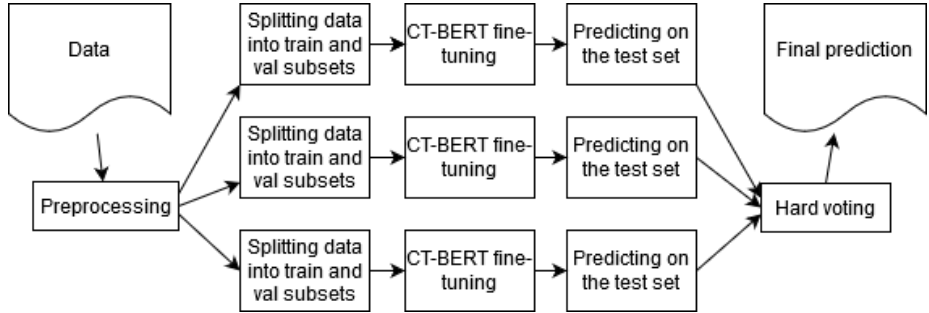}}
\caption{Our approach to COVID-19 fake news detection.}
\label{ris:image}
\end{figure}

Four of our five submitted models were trained entirely on the dataset provided by the competition organizers \cite{dataset}. The last model was trained on the competition data and on additional datasets \cite{add_data2,add_data1}. The training details and the results of our final submissions are summarised in Table 3.

\begin{longtable}{|p{0.15\linewidth}|p{0.17\linewidth}|p{0.13\linewidth}|p{0.45\linewidth}|}
\caption{Final submissions}
\\
\hline
Place among all submissions & Submission name & Weighted F1-score (\%) & Training details \\ \hline
1 & g2tmn\_2.csv & 98.69 & All models were trained both on training and validation sets with no hold out validation. We trained 3 models and then used hard-voting to ensemble their predictions together. \\ \hline
6 & g2tmn\_4.csv & 98.51 & 1000 random posts were used for hold-out validation. Models were trained on all other data. We trained 5 models with random seed values and choose 3 models with the best F1-score performances to ensemble them together. \\ \hline
11 & g2tmn\_1.csv & 98.37 & Models were trained on the official training set. The validation set was used for hold-out validation. We trained 5 models with random seed values and choose 3 best-performance models to ensemble their predictions. \\ \hline
15 & g2tmn\_3.csv & 98.32 & This submission was similar to g2tmn\_1.csv but it had different seed values. \\ \hline
25 & g2tmn\_5.csv & 98.18 & 1000 random posts were used for hold-out validation. Models were trained on all other official data, CoAID and FakeCovid data. We trained 5 models with random seed values and used 3 best-performance models for ensembling learning.\\ \hline
\end{longtable}

It can be seen from the data in Table 3 that, with the weighted F1-score, our model performance is 98.69\% (the random seeds are 23, 30, and 42), which was ranked the first place of the leaderboard of this task.

\subsection{Error Analysis}

Error analysis allows us to further evaluate the quality of the machine learning model and conduct a quantitative analysis of errors. Figure 2 provides the confusion matrix for our best solution when detecting fake news about COVID-19 on the test set. As can be seen from the figure, the precision of our system is slightly higher than its recall. In other words, the false positive value is greater that false negative. Table 4 shows the examples of false negative and false positive errors. 

We noticed that the type of error is frequently related to the topic of the post. For example, the model often misclassifies false reports about the number of people infected. At the same time, true posts related to the coronavirus vaccine or to political topics can be identified as false.

\begin{figure}[h]
\center{\includegraphics[width=0.8\linewidth]{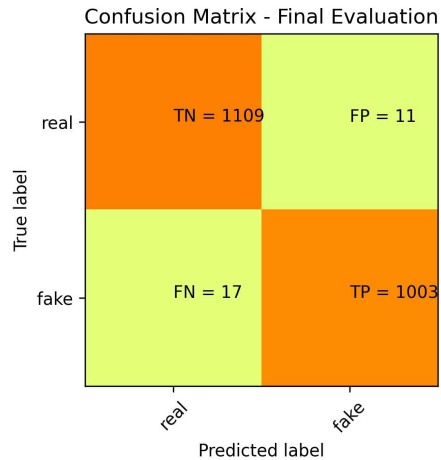}}
\caption{Confusion matrix of our best-performance ensemble for COVID-19 fake news detection (for the fake class).}
\label{ris:image}
\end{figure}

\begin{longtable}{|p{0.15\linewidth}|p{0.15\linewidth}|p{0.65\linewidth}|}
\caption{Some examples of misclassified posts}
\\
\hline
True label & Prediction & Post \\ \hline
real & fake & Scientists ask: Without trial data how can we trust Russia's \#COVID vaccine? https://t.co/gArcUf0Pji https://t.co/0bdcA7lf56 \\ \hline
real & fake & *DNA Vaccine: injecting genetic material into the host so that host cells create proteins that are similar to those in the virus against which the host then creates antibodies \\ \hline
real & fake & Donald Trump has claimed he "up-played" the seriousness of the coronavirus pandemic - despite admitting earlier this year he had "wanted to always play it down" https://t.co/wEgnnZzrNW \\ \hline
fake & real & Govt has added \#Corona disease in all existing mediclaim insurances as a special case \#COVID2019India https://t.co/39vpW7tBqq \\ \hline
fake & real & As tuberculosis shaped modernism, so COVID-19 and our collective experience of staying inside for months on end will influence architecture's near future, @chaykak writes. https://t.co/ag34yZckbU \\ \hline
fake & real & Northern Ireland was testing for COVID-19 at a rate 10 times that of Scotland reported on 9 May 2020. \\ \hline
\end{longtable}

\section{Conclusion}

In this work, we propose a simple but effective approach to COVID-19 fake news detection based on CT-BERT and ensembling learning. Our experiments confirmed that BERT-based models specialized in the subject area successfully cope with such tasks and perform high-quality binary classification.

The experimental results showed that our solution achieved 98.69\% of the weighted F1-score on test data and ranked in the first place in the Constraint@-AAAI2021 shared task. For future work, we can experiment with different training and data augmentation techniques. We can also apply and evaluate hybrid models combining BERT-based architectures with other methods of natural language processing \cite{further1,further2}. 
%
%
%
%

\end{document}